\documentclass[letterpaper]{article} 
\usepackage{aaai24}  
\usepackage{times}  
\usepackage{helvet}  
\usepackage{courier}  
\usepackage[hyphens]{url}  
\usepackage{graphicx} 
\usepackage{natbib}  
\usepackage{caption} 
\usepackage{algorithm}
\usepackage{algorithmic}

\usepackage{graphicx}
\usepackage{amsmath}
\usepackage{amssymb}
\usepackage{booktabs}

\usepackage{url}            
\usepackage{amsfonts}       
\usepackage{microtype}      
\usepackage{xcolor}         
\usepackage{multirow}
\usepackage{colortbl}
\usepackage{enumitem}
\usepackage{subcaption}

\usepackage{newfloat}
\usepackage{listings}
\DeclareCaptionStyle{ruled}{labelfont=normalfont,labelsep=colon,strut=off} 
\floatstyle{ruled}
\newfloat{listing}{tb}{lst}{}
\floatname{listing}{Listing}
\title{Decoupled Contrastive Learning for Long-Tailed Recognition}
\author{
    Shiyu~Xuan,
    Shiliang~Zhang
}
\affiliations{
    National Key Laboratory for Multimedia Information Processing\\
    School of Computer Science, Peking University, Beijing, China \\
    shiyu\_xuan@stu.pku.edu.cn, slzhang.jdl@pku.edu.cn
}

\usepackage{bibentry}

\begin{document}
\frenchspacing
\maketitle

\begin{abstract}
    Supervised Contrastive Loss (SCL) is popular in visual representation learning.
    Given an anchor image, SCL pulls two types of positive samples, \emph{i.e.}, its augmentation and other images from the same class together, while pushes negative images apart to optimize the learned embedding. In the scenario of long-tailed recognition, where the number of samples in each class is imbalanced, treating two types of positive samples equally leads to the biased optimization for intra-category distance. In addition, similarity relationship among negative samples, that are ignored by SCL, also presents meaningful semantic cues. To improve the performance on long-tailed recognition, this paper addresses those two issues of SCL by decoupling the training objective. Specifically, it decouples two types of positives in SCL and optimizes their relations toward different objectives to alleviate the influence of the imbalanced dataset. We further propose a patch-based self distillation to transfer knowledge from head to tail classes to relieve the under-representation of tail classes. It uses patch-based features to mine shared visual patterns among different instances and leverages a self distillation procedure to transfer such knowledge. Experiments on different long-tailed classification benchmarks demonstrate the superiority of our method. For instance, it achieves the 57.7\% top-1 accuracy on the ImageNet-LT dataset. Combined with the ensemble-based method, the performance can be further boosted to 59.7\%, which substantially outperforms many recent works. The code is available at \url{https://github.com/SY-Xuan/DSCL}.
 \end{abstract}
 
 \section{Introduction}
 \begin{figure}[t!]
     \centering
     \includegraphics[width=0.87\linewidth]{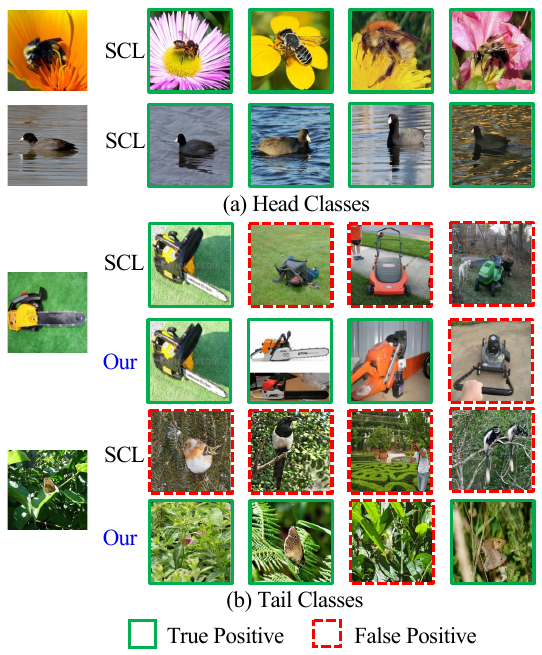}
     \caption{Examples of retrieval results using features learned by SCL on head classes in (a) and tail classes in (b). In (b), features learned by SCL are biased to low-level appearance cues, while features learned by our method are more discriminative to semantic cues.}
    \label{fig:motivation}
   \end{figure}
 
 Thanks to the powerful deep learning methods, the performance of various vision tasks~\cite{Russakovsky2015,long2015fully} on manually balanced dataset has been significantly boosted. In real-world applications, training samples commonly exhibit a long-tailed distribution, where a few head classes contribute most of the observations, while lots of tail classes are associated with only a few samples~\cite{van2018inaturalist}. Long-tailed distribution leads to two challenges for visual recognition: (a) the loss function designed for the balanced dataset can be easily biased toward the head classes. (b) each of tail classes contains too few samples to represent visual variances, leading to the under-representation of the tail classes.
 
 By optimizing the intra-inter category distance, Supervised Contrastive Loss (SCL)~\cite{Khosla2020} has achieved impressive performance on balanced datasets. Given an anchor image, SCL pulls two kinds of positive samples together, \emph{i.e.}, (a) different views of the anchor image generated by the data augmentation, and (b) other images from the same classes. Those two types of positives supervise the model to learn different representations, \emph{i.e.}, images from the same categories enforce the learning of semantic cues, while samples augmented by appearance variances mostly lead to the learning of low-level appearance cues. Fig.~\ref{fig:motivation} (a) shows that, SCL effective learns semantic features for head classes, \emph{e.g.},the learned semantic ``bee'' is robust to cluttered backgrounds. As shown in Fig.~\ref{fig:motivation} (b), representations learned by SCL for tail classes are more discriminative to low-level appearance cues like shape, texture, and color.
 
 Our theoretical analysis in Section {Overview} indicates that, SCL poses imbalanced gradients on two kinds of positive samples, resulting a biased optimization for head and tail classes. We hence proposes the Decoupled Contrastive Learning, which adopts the Decoupled Supervised Contrastive Loss (DSCL) to handle this issue. Specifically, DSCL decouples two kinds of positive samples to re-formulate the optimization of intra-category distance. It alleviates the imbalanced gradients of two kinds of positive samples. We also give a theoretical proof that DSCL prevents the learning of a biased intra-category distance. In Fig.~\ref{fig:motivation} (b), features learned by our method are discriminative to semantic cues, and substantially boost the retrieval performance on tail classes.
 
 To further alleviate the challenge of long-tailed distribution, we propose the Patch-based Self Distillation (PBSD) to leverage head classes to facilitate the representation learning in tail classes. PBSD adopts a self distillation strategy to better optimize the inter-category distance, through mining shared visual patterns between different classes and transferring knowledge from head to tail classes. We introduce patch-based features to represent visual patterns from an object. The similarity between patch-based features and instance-level features is calculated to mine the shared visual patterns, \emph{i.e.}, if an instance shares visual patterns with a patch-based feature, they will have high similarity. We leverage the self distillation loss to maintain the similarity relationship among samples, and integrate the knowledge into the training.
 
 DSCL and PBSD are easy to implement, and substantially boosts the long-tailed recognition performance. We evaluate our method on several long-tailed datasets including ImageNet-LT~\cite{liu2019large}, iNaturaList 2018~\cite{van2018inaturalist}, and Places-LT~\cite{liu2019large}. Experimental results show that our method improves the SCL by 6.5\% and achieves superior performance compared with recent works. For example, it outperforms a recent contrastive learning based method TSC~\cite{li2021targeted} by 5.3\% on ImageNet-LT. Our method can be flexibly combined with ensemble-based methods like RIDE~\cite{wang2020long}, which achieves the overall accuracy of 74.9\% on the iNaturaList 2018, outperforming the recent work CR~\cite{ma2023curvature} by 1.4\% in overall accuracy.
 
 To the best of our knowledge, this is an original contribution decoupling two kinds of positives and using patch-based self distillation to boost the performance of SCL on long-tail recognition. The proposed DSCL decouples different types of positive samples to pursue a more balanced intra-category distance optimization across head and tail classes. It also introduces the similarity relationship cues to leverage shared patterns in head classes for the optimization in tail classes. Extensive experiments on three commonly used datasets have shown its promising performance. Our method is easy to implement and the code will be released to benefit the future research on long-tailed visual recognition.

 \section{Related Work}
 
 \textbf{Long-tailed recognition} aims to address the problem of the model training in the situation where a small portion of classes have massive samples but the others are associated with only a few samples. Current research can be summarized into four categories, \emph{e.g.}, re-balancing methods, decoupling methods, transfer learning methods and ensemble-based methods, respectively.

 Re-balancing methods use re-sampling or re-weighting to deal with long-tailed recognition. Re-sampling methods typically include over-sampling for the tail classes~\cite{byrd2019effect} or under-sampling for the head classes~\cite{japkowicz2002class}.
 Besides re-sampling, re-weighting the loss function is also an effective solution.
 For instance, Balanced-Softmax~\cite{ren2020balanced} presents the unbiased extension of Softmax based on the Bayesian estimation.
 Re-balancing methods could be harmful to the discriminative power of the learned backbone~\cite{kang2019decoupling}. Therefore, decoupling methods propose the two-stage training to decouple the representation learning and the classifier training.
 Transfer learning methods enhance the performance of the model by transferring knowledge from head classes to tail classes.
 BatchFormer~\cite{hou2022batchformer} introduces a one-layer Transformer~\cite{vaswani2017attention} to transfer knowledge by learning the sample relationship from each mini-batch.
 Ensemble-based methods leverage multi experts to solve long-tailed visual learning. RIDE~\cite{wang2020long} proposes a multi-branch network to learn diverse classifiers in parallel.
 Although ensemble-based method achieves superior performance, the introduction of multi experts increases the number of parameters and computational complexity.
 
 \textbf{Contrastive learning} has received much attention because of its superior performance on representation learning~\cite{He2020}. Contrastive learning aims to find a feature space that can encode the semantic similarities by pulling the positive pairs together while pushing negative pairs apart. Some researchers have leveraged contrastive learning in the long-tailed recognition. For example, KCL~\cite{kang2020exploring} finds that the self-supervised learning based on contrastive learning can learn a balanced feature space. To leverage the useful label information, they extend SCL~\cite{Khosla2020} by introducing a k-positive sampling method. TSC~\cite{li2021targeted} improves the uniformity of the feature distribution by making features of different classes converge to a pre-defined uniformly distributed targets. Some methods~\cite{yun2022patch,zhang2023patch} extend contrastive learning with localized information to benefit the dense prediction tasks.
 
 This work differs with previous ones in several aspects. Existing long-tailed recognition works using contrastive learning treat two kinds of positives equally. To the best of our knowledge, this is an early work revealing that treating two kinds of positives equally leads to a biased optimization across categories. We hence propose a decoupled supervised contrastive loss to pursue a balanced intra-category distance optimization. We further extend the contrastive learning by introducing patch-based self distillation to transfer knowledge between classes, mitigating the under-representation of the tailed classes and leading to a more effective optimization to inter-category distance. Different from other transfer learning methods, PBSD leverages patch-based features to mine shared patterns between different classes and designs a self distillation procedure to transfer knowledge. The self distillation procedure does not rely on a large teacher model or multi-expert models~\cite{li2022nested}, making it efficient. 
 Compared with patch-based contrastive learning methods that only mine similar patches from different views of an image, PBSD transfers knowledge between different images.
 Those differences and the promising performance in extensive experiments highlight the contribution of this work.
 
 \section{Methodology}~\label{sec:method}
 \begin{figure}[t!]
    \centering
     \includegraphics[width=0.9\linewidth]{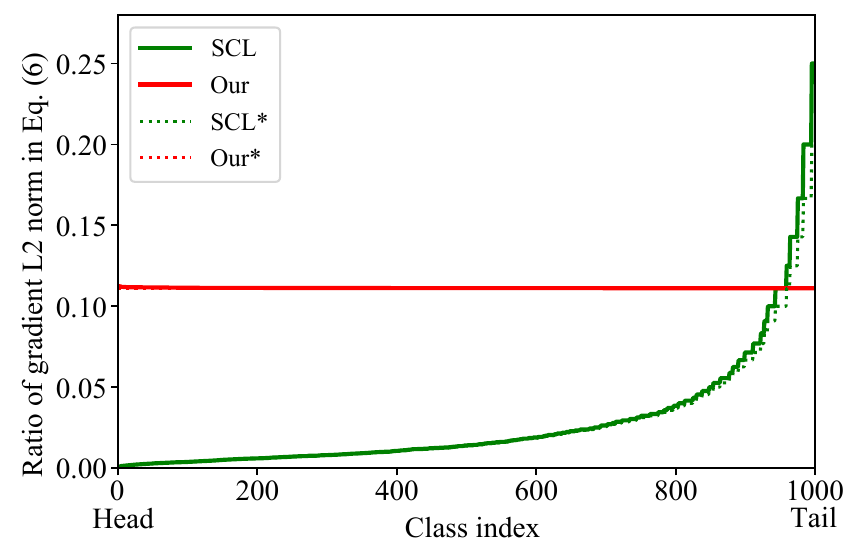}
    \caption{Average ratio of gradient L2 norm computed by pulling the anchor with two types of positives as in Eq.~\eqref{eq:gradient_ratio} on ImageNet-LT. `*' denotes the theoretical ratio. SCL treats two types of positives equally, and leads to the imbalanced optimization. Two types of positives denote the data argumentation and other images in the same category.}
    \label{fig:gradient_ratio}
   \end{figure}
 
 \begin{figure*}[t!]
    \centering
     \includegraphics[width=0.85\linewidth]{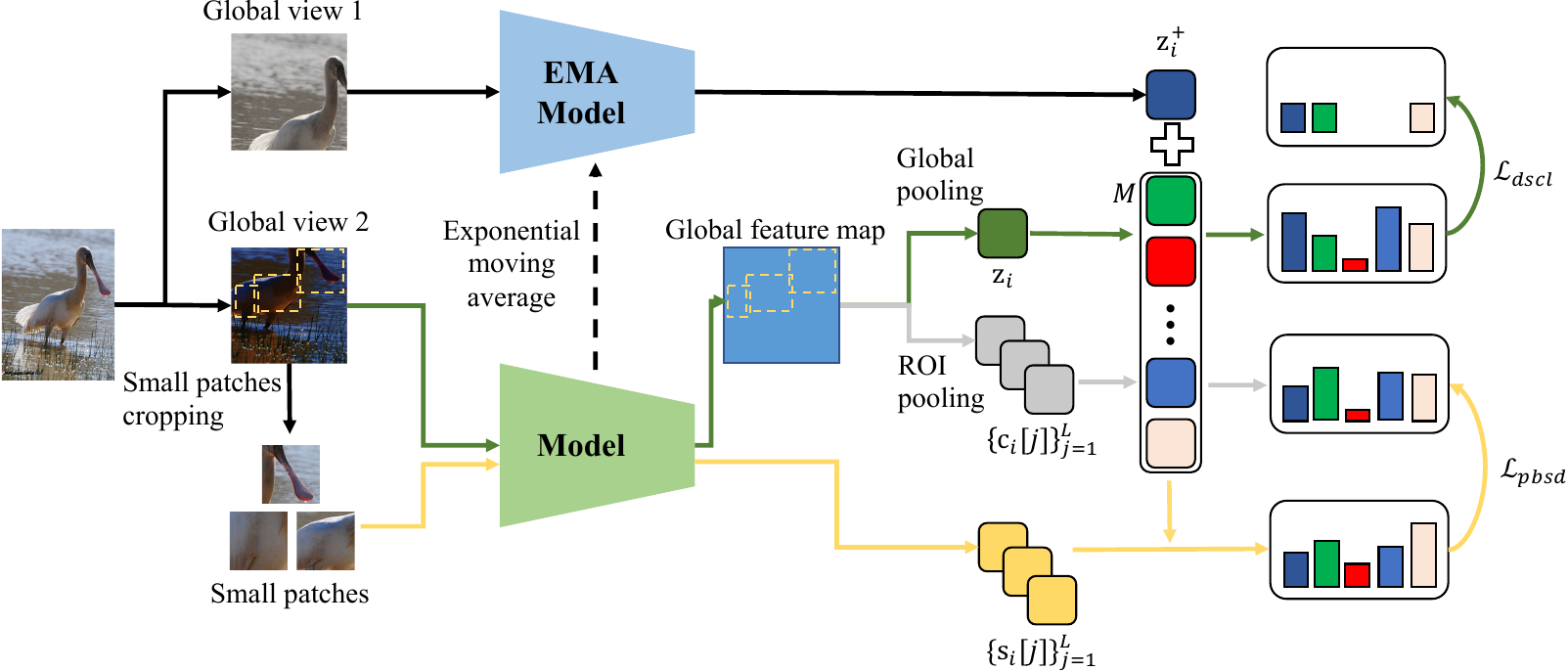}
     \caption{Illustration of the proposed method. Data augmentation is performed to get two global views of a training image. Then small patch is cropped from the global view. The backbone and Exponential Moving Average (EMA) backbone~\cite{He2020} are used to extract normalized features. These features are used to calculate the similarity distribution with memory queue $M$. $\mathcal{L}_{dscl}$ optimizes the feature space by pulling the anchor image with its positive samples together and pushing the anchor image with its negative samples apart. $\mathcal{L}_{pbsd}$ transfers knowledge through mimicking two similarity distributions.
     }
     \label{fig:framework}
 \end{figure*}
 
 \subsection{Analysis of SCL}\label{sec:overview}
 Given a training dataset $\mathcal{D}=\{\mathrm{x}_i, \mathrm{y}_i\}_{i=1}^n$, where $\mathrm{x}_i$ denotes an image and $\mathrm{y}_i \in \{1,\dots,K\}$ is its class label. Assuming that $n^k$ denotes the cardinality of class $k$ in $\mathcal{D}$, and indexes of classes are sorted by cardinality in decreasing order, \emph{i.e.}, if $a< b$, then $n^{a} \ge n^{b}$. In long-tailed recognition, the training dataset is imbalanced, \emph{i.e.}, $n^1 \gg n^K$, and the imbalance ratio is defined as $n^1 / n^{K}$. For the image classification task, algorithm aims to learn a feature extraction backbone $\mathrm{v}_i=\operatorname f_{\theta}(\mathrm{x}_i)$, and a linear classifier, which first maps an image $\mathrm{x}_i$ into a global feature map $\mathrm{u}_i$ and uses the global pooling to get a $d$-dim feature vector $\mathrm{v}_i$. It hence classifies the feature vector to a $K$-dim classification score. Typically, the testing dataset $\mathcal{T}$ is balanced.
 
 Supervised Contrastive Learning (SCL) is commonly adopted to learn the feature extraction backbone. Given an anchor image $\mathrm{x}_{i}$, defining $\mathrm{z}_{i}=\operatorname{g}_{\gamma}(\mathrm{v}_i)$ as the normalized feature extracted with the backbone and an extra projection head $\operatorname{g}_{\gamma}$~\cite{He2020}, $\mathrm{z}_{i}^{+}$ as the normalized feature of a positive sample of $\mathrm{x}_{i}$ generated by data augmentation. We use $M$ to denote a set of sample features that can be acquired by the memory queue~\cite{He2020}, and use $P_i$ as the positive feature set of $\mathrm{x}_{i}$ drawn from $M$ with $P_i = \{ \mathrm{z}_{t} \in M: \mathrm{y}_{t}=\mathrm{y}_{i}\}$. SCL decreases the intra-category distance by pulling the anchor image and its positive samples together, meanwhile enlarges the inter-category distance through pushing images with different class labels apart, \emph{i.e.},
 
 \begin{equation}
    \small
 \mathcal{L}_{scl} = \frac{-1}{|P_i| + 1} \sum_{\mathrm{z}_t \in \{\mathrm{z}_i^{+} \cup P_i\}} \log p(\mathrm{z}_t |\mathrm{z}_i),
 \label{eq:loss_SL}
 \end{equation}
 where $|P_i|$ is the cardinality of $P_i$. Using $\tau$ to denote a predefined temperature parameter, the conditional probability $p(\mathrm{z}_t |\mathrm{z}_i)$ is computed as,
 \begin{equation}
    \small
 p(\mathrm{z}_t |\mathrm{z}_i) = \frac{\exp (\mathrm{z}_t \cdot \mathrm{z}_i / \tau)}{\sum\limits_{\mathrm{z}_m \in \{\mathrm{z}_i^{+} \cup M\}} \exp (\mathrm{z}_m \cdot \mathrm{z}_{i} / \tau)}.
 \label{eq:conditional distribution}
 \end{equation}
 
 Eq.~\eqref{eq:loss_SL} can be formulated as a distribution alignment task,
 \begin{equation}
    \small
   \mathcal{L}_{align} = \sum_{\mathrm{z}_t \in \{\mathrm{z}_i^{+} \cup M\}} - \hat{p}(\mathrm{z}_t |\mathrm{z}_i) \log p(\mathrm{z}_t |\mathrm{z}_i),
   \label{eq:unify}
 \end{equation}
 where $\hat{p}(\mathrm{z}_t|\mathrm{z}_i)$ is the probability of the target distribution. For $\mathrm{z}_{i}^{+}$ and $\mathrm{z}_t \in P_i$, SCL treats them equally as positive samples and sets their target probability as $1 / (|P_i| + 1)$. For other images with different class labels in $M$, SCL treats them as negative samples and sets their target probability as 0.
 
 For the feature $\mathrm{z}_i$ of an anchor image $\mathrm{x}_i$, the gradient of SCL is,
 \begin{equation}
    \small
     \begin{aligned}
     \frac{\partial \mathcal{L}_{scl}}{\partial \mathrm{z}_{i}} = \frac{1}{\tau} &\Bigg\{\sum_{\mathrm{z}_j \in N_i} \mathrm{z}_j p(\mathrm{z}_j|\mathrm{z}_i)
     ~ + \mathrm{z}_i^{+}\left(p(\mathrm{z}_i^{+}|\mathrm{z}_i) - \frac{1}{|P_i| + 1}\right) \\
     & + \sum_{\mathrm{z}_t \in P_i} \mathrm{z}_t \left(p(\mathrm{z}_t|\mathrm{z}_i) - \frac{1}{|P_i| + 1}\right) \Bigg\}, \\
     \end{aligned}
     \label{eq:gradient_SCL}
 \end{equation}
 where $N_i$ is the negative set of $\mathrm{x}_{i}$ containing features drawn from $\{\mathrm{z}_{j} \in M: \mathrm{y}_{j} \neq \mathrm{y}_{i}\}$.
 SCL involves two types of positive samples $\mathrm{z}_i^+$ and $\mathrm{z}_t \in P_i$.
 We compute gradients of pulling the anchor with two types of positive samples as,
 \begin{equation}
    \small
     \begin{aligned}
     & \left.\frac{\partial \mathcal{L}_{scl}}{\partial \mathrm{z}_{i}}\right\vert_{\mathrm{z}_i^+} = \mathrm{z}_i^{+}\left(p(\mathrm{z}_i^{+}\mid \mathrm{z}_i) - \frac{1}{|P_i| + 1}\right), \\
     & \left.\frac{\partial \mathcal{L}_{scl}}{\partial \mathrm{z}_{i}}\right\vert_{\mathrm{z}_t} = \mathrm{z}_t \left(p(\mathrm{z}_t|\mathrm{z}_i) - \frac{1}{|P_i| + 1}\right), \mathrm{z}_t \in P_i. \\
     \end{aligned}
 \end{equation}
 
 At the beginning of the training, the ratio of gradient L2 norm of two kinds of positive samples is,
 \begin{equation}
    \small
     \frac{\left\Vert \left.\frac{\partial \mathcal{L}_{scl}}{\partial \mathrm{z}_{i}}\right\vert_{\mathrm{z}_i^+}\right\Vert_2 }{\sum\limits_{\mathrm{z}_t \in P_i}\left\Vert \left.\frac{\partial \mathcal{L}_{scl}}{\partial \mathrm{z}_{i}}\right\vert_{\mathrm{z}_t}\right\Vert_2} \approx \frac{1}{|P_i|}.
     \label{eq:gradient_ratio}
 \end{equation}
 
 When SCL converges, the optimal conditional probability of $\mathrm{z}_i^{+}$ is,
 \begin{equation} ~\label{eq:converge_p}
    \small
     p(\mathrm{z}_i^{+}|\mathrm{z}_i) = \frac{1}{|P_i| + 1}.
 \end{equation}
 A detailed proof of above computations can be found in the Supplementary Material.
 
 In SCL, the memory queue $M$ is uniformly sampled from the training set, which leads to $|P_i| \approx \frac{n^{\mathrm{y}_i}}{n} |M|$. In a balanced dataset, $n^1 \approx n^2 \approx \dots \approx n^K$, resulting a balanced $|P_i|$ across different categories. For a long-tail dataset with imbalanced $|P_i|$, SCL makes the head classes pay more attention to pulling the anchor $z_i$ with features from $P_i$ together as the gradient is dominated by the third term in Eq.~\eqref{eq:gradient_SCL}.
 
 As shown in Fig.~\ref{fig:gradient_ratio}, the ratio of gradient L2 norm of pulling two kinds of positive samples is unbalanced.
 When the training of SCL converges, the optimal value of $p(\mathrm{z}_i^{+}|\mathrm{z}_i)$ is also influenced by the $|P_i|$ as shown in Eq.~\eqref{eq:converge_p}. The inconsistency of learned features across categories is illustrated in Fig.~\ref{fig:motivation}(a) and (b). This phenomena has also been validated by~\cite{wei2020can} that pulling $z_i$ with ${z}_i^{+}$ and samples from $P_i$ leads to learning different representations, \emph{i.e.}, appearance features for tail classes and semantic features for head classes, respectively.
 
 Eq.~\eqref{eq:gradient_SCL} also indicates that, SCL equally pushes away all the negative samples to enlarges the inter-category distance. This strategy ignores the valuable similarity cues among different classes.
 To seek a better way to optimize intra and inter category distance, we propose the Decoupled Supervised Contrastive Loss (DSCL) to decouple two kinds of positive samples to prevent the biased optimization, as well as the Patch-based Self Distillation (PBSD) to leverage similarity cues among classes.
 
 \subsection{Decoupled Supervised Contrastive Loss}

 DSCL is proposed to ensure a more balanced optimization to the intra-category distance across different categories. It decouples two kinds of positive samples and add different weights on them to make the gradient L2 norm ratio and the optimal value of $p(\mathrm{z}_i^{+}|\mathrm{z}_i)$ not influenced by the number of samples in each category. We represent the DSCL as,
 \begin{equation}
    \small
     \mathcal{L}_{dscl} = \frac{-1}{|P_i| + 1} \sum_{\mathrm{z}_t \in \{\mathrm{z}_i^{+} \cup P_i\}} \log \frac{\exp w_t (\mathrm{z}_t \cdot \mathrm{z}_{i} / \tau)}{\sum\limits_{\mathrm{z}_m \in \{\mathrm{z}_i^{+} \cup M\}} \exp (\mathrm{z}_m \cdot \mathrm{z}_{i} / \tau)},
     \label{eq:decouple_target2}
 \end{equation}
 where,
 \begin{equation}
    \small
     w_t = \left\{\begin{aligned}
         &\alpha (|P_i| + 1) , &&\mathrm{z}_t = \mathrm{z}_i^{+} \\
         &\frac{(1 - \alpha)(|P_i| + 1)}{|P_i|} , &&\mathrm{z}_t \in P_i \\
     \end{aligned}\right.
 \label{eq:decouple_target}
 \end{equation}
 where $\alpha \in [0, 1]$ is a pre-defined hyper-parameter. The proposed DSCL is a generalization of SCL in both balanced setting and imbalanced setting. If the dataset is balanced, DSCL is the same as SCL by setting $\alpha=1 / (|P_i| + 1)$.
 
 We proceed to show why Eq.~\eqref{eq:decouple_target2} leads to a more balanced optimization.
 
 {At the beginning of the training, the gradient L2 norm ratio of two kinds of positives is,}
 \begin{equation}
    \small
     \frac{\left\Vert \left.\frac{\partial \mathcal{L}_{dscl}}{\partial \mathrm{z}_{i}}\right\vert_{\mathrm{z}_i^+}\right\Vert_2 }{\sum\limits_{\mathrm{z}_t \in P_i}\left\Vert \left.\frac{\partial \mathcal{L}_{dscl}}{\partial \mathrm{z}_{i}}\right\vert_{\mathrm{z}_t}\right\Vert_2} \approx \frac{\alpha}{1 - \alpha}.
     \label{eq:gradient_ratio_DSCL}
 \end{equation}
 {When DSCL converges, the optimal conditional probability of $\mathrm{z}_i^{+}$ is $p(\mathrm{z}_i^{+}|\mathrm{z}_i) = \alpha$,}
 where a detailed proof can be found in the Supplementary Material.
 
 As shown in Eq.~\eqref{eq:gradient_ratio_DSCL} and Fig.~\ref{fig:gradient_ratio}, the gradient ratio of two kinds of positive samples is not influenced by $|P_i|$. DSCL also ensures that the optimal value of $p(\mathrm{z}_i^{+}|\mathrm{z}_i)$ is not influenced by $|P_i|$, hence alleviates the inconsistent feature learning issue between head and tail classes.
 
 \subsection{Patch-based Self Distillation}

\begin{figure}[t!]
    \centering
   \includegraphics[width=1\linewidth]{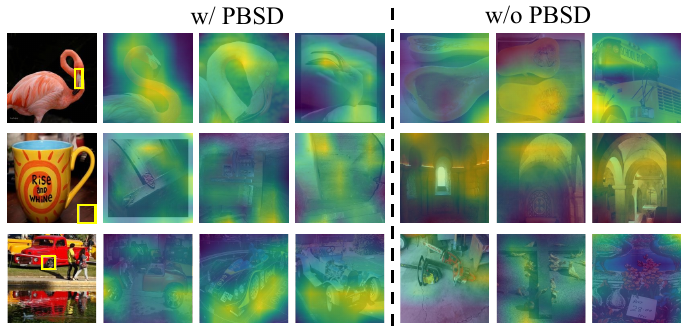}
   \caption{Patch-based image retrieval results (top 3 returned) on ImageNet-LT. Query patches are highlighted with yellow bounding boxes. The response map of query patch features on the retrieved images is also illustrated.}
   \label{fig:patch_vis}
 \end{figure}

 Visual patterns can be shared among different classes, \emph{e.g.}, the visual pattern `wheel' is shared by the class `truck', `car', and `bus'. Features of many visual patterns in tail classes can be learned from head classes that share these visual patterns, hence reducing the difficult of representation learning in tail classes. SCL pushes two instances from different classes away in the feature space, even they share meaningful visual patterns. As shown in Fig.~\ref{fig:patch_vis}, we extract query patch features from yellow bounding boxes and retrieve the top-3 similar samples from the dataset. Retrieval results of SCL denoted by `w/o PBSD' are not semantically related to the query patch, indicating that SCL is not effective in learning and leveraging patch-level semantic cues.
 
 Inspired by the patch-based methods in fine-grained image recognition~\cite{zhang2014part,quan2019auto,sun2018beyond}, we introduce patch-based features to encode visual patterns. Given the global feature map $\mathrm{u}_i$ of an image $\mathrm{x}_i$ extracted by the backbone, we first randomly generate some patch boxes.
 The coordinates of those patch boxes denote $\{\mathrm{B}_i[j]\}_{j=1}^L$, where $L$ is the number of the patch boxes.
 We hence apply ROI pooling~\cite{he2017mask} according to the coordinates of those patch boxes and send pooled features into a projection head to get the normalized embedding features $\{\mathrm{c}_i[j]\}_{j=1}^L$ with
 \begin{equation}
    \small
     \mathrm{c}_i[j] = \operatorname{g}_{\gamma}\left(\operatorname{ROI}\left(\mathrm{u}_i, \mathrm{B}_i[j]\right)\right).
 \label{eq:roi_pooling}
 \end{equation}
 
 Similar to Eq.~\eqref{eq:conditional distribution}, the conditional probability is leveraged to calculate the similarity relationship between instances,
 \begin{equation}
    \small
     p(\mathrm{z}_t |\mathrm{c}_i^j) = \frac{\exp (\mathrm{z}_t \cdot \mathrm{c}_{i}[j] / \tau)}{\sum\limits_{\mathrm{z}_m \in \{\mathrm{z}_i^+ \cup M\}} \exp (\mathrm{z}_m \cdot \mathrm{c}_{i}[j] / \tau)}.
 \label{eq:target_pbsd}
 \end{equation}
 If the image corresponding to $\mathrm{z}_t$ has shared visual patterns with the patch-based features, $\mathrm{z}_t$ and $\mathrm{c}_{i}[j]$ will have a high similarity. Therefore, Eq.~\eqref{eq:target_pbsd} encodes the similarity cues between each pair of instances.
 
 We use the similarity cues as the knowledge to supervise the training procedure. To maintain such knowledge, we also crop multi image patches from the image according to $\{\mathrm{B}_i[j]\}_{j=1}^L$, and extract their feature embeddings $\{\mathrm{s}_{i}[j]\}_{j=1}^L$ with the backbone,
 \begin{equation}
    \small
     \mathrm{s}_i[j] = \operatorname{g}_{\gamma}\left(\operatorname{f}_{\theta}\left(\operatorname{Crop}(\mathrm{x}_i, B_i[j])\right)\right).
 \label{eq:patch_crop}
 \end{equation}
 
 PBSD enforces the feature embeddings of image patches to produce the same similarity distribution as the patch-based features via the following loss,
 \begin{equation}
    \small
 \mathcal{L}_{pbsd} = \frac{1}{L} \sum_{j=1}^L \sum_{\mathrm{z}_t \in \{\mathrm{z}_{i}^{+} \cup M\}} - p(\mathrm{z}_t|\mathrm{c}_i[j]) \log p(\mathrm{z}_t|\mathrm{s}_{i}[j]), \\
 \label{eq:loss_pbsd}
 \end{equation}
 note that $p(\mathrm{z}_t|\mathrm{c}_i[j])$ is detached from the computation graph to block the gradient.
 
 Local visual patterns of an object can be shared by different categories. We hence use patch-based features to represent visual patterns. $p(\mathrm{z}_t |\mathrm{c}_i[j])$ is calculated to mine relationship of the shared patterns among images. Minimizing Eq.~\eqref{eq:loss_pbsd} maintains shared patterns to transfer knowledge and mitigate the under-representation of the tailed classes. The retrieval results shown in Fig.~\ref{fig:patch_vis} indicate that our method effectively reinforces the learning of patch-level features and patch-to-image similarity, making it possible to mine shared visual patterns of different classes.
 Experiments also validate that PBSD loss is important to the performance gain.
 
 Multi-crop trick~\cite{caron2020unsupervised} is commonly used in self-supervised learning to generate more augmented samples of the anchor image. It introduces low resolution crops to reduce the computational complexity. Our motivation and loss design are different with the multi-crop strategy. PBSD is motivated to leverage shared patterns between head and tail classes to assist the learning of the tail classes. Patch-based features are obtained with ROI pooling to represent shared patterns. Eq.~\eqref{eq:loss_pbsd} performs self distillation to maintain shared patterns. We conduct an experiment by replacing PBSD with the multi-crop trick. As shown in Table~\ref{tab:components}, the performance on ImageNet-LT drops from 57.7\% to 56.1\%, indicating that PBSD is more effective than the multi-crop strategy.
 
 \subsection{Training Pipeline}
 We illustrate our method in Fig.~\ref{fig:framework}. To maintain a memory queue, we use a momentum updated model as in~\cite{He2020}. The training is supervised by two losses, \emph{i.e.}, the decoupled supervised contrastive loss and the patch-based self distillation loss. The overall training loss is denoted as,
 \begin{equation}
    \small
 \mathcal{L}_{overall} = \mathcal{L}_{dscl} + \lambda \mathcal{L}_{pbsd},
 \label{eq:loss_overall}
 \end{equation}
 where $\lambda$ is the loss weight.
 
 Our method focuses on the representation learning, and can be applied in different tasks by concatenating their losses. Following~\cite{li2021targeted,kang2020exploring}, after the training of the backbone, we discard the learned projection head $\operatorname{g}_{\gamma}(\cdot)$ and train a linear classifier on top of the learned backbone using the standard cross-entropy loss with the class-balanced sampling strategy~\cite{kang2019decoupling}. The following section proceeds to present our evaluation to the proposed methods.
 
 \section{Experiments}
 \subsection{Experimental Setup}
 \textbf{Datasets.} We use three popular datasets to evaluate the long-tailed recognition performance.
 \begin{itemize}
 \item \textbf{ImageNet-LT}~\cite{liu2019large} contains 115,846 training images of 1,000 classes sampled from the ImageNet1K~\cite{Russakovsky2015}, with class cardinality ranging from 5 to 1,280.
 \item \textbf{iNaturaList 2018}~\cite{van2018inaturalist} is a real-world long-tailed dataset with 437,513 training images of 8,142 classes, with class cardinality ranging from 2 to 1,000.
 \item \textbf{Places-LT}~\cite{liu2019large} contains 62,500 training images of 365 classes sampled from the Places~\cite{zhou2017places}, with class cardinality ranging from 5 to 4,980.
 \end{itemize}
 
 \noindent\textbf{Evaluation Metrics.} We follow the standard evaluation metrics that evaluate our models on the testing set and report the overall top-1 accuracy across all classes. To give a detailed analysis, we follow~\cite{liu2019large} that groups the classes into splits according to their number of images: Many ($>$ 100 ), Medium (20 - 100), and Few ($<$ 20).
 
 \noindent\textbf{Implementation Details.}
 For a fair comparison, we follow the implementations of TSC~\cite{li2021targeted} and KCL~\cite{kang2020exploring} that train the backbone at the first stage and train the linear classifier with the freezed learned backbone at the second stage.
 We adopt ResNet-50~\cite{He2016} as backbone for all experiments except that using ResNet-152 pre-trained on ImageNet1K for Places-LT. The $\alpha$ in Eq.~\eqref{eq:decouple_target} is set as 0.1 and the loss weight $\lambda$ in Eq.~\eqref{eq:loss_overall} is 1.5.
 
 At the first stage, the basic framework is the same as MoCoV2~\cite{Chen2020}, the momentum value for the updating of EMA model is 0.999, the temperature $\tau$ is 0.07, the size of memory queue $M$ is 65536, and the output dimension of projection head is 128. The data augmentation is the same as MoCoV2~\cite{Chen2020}. Locations to get the patch-based features are sampled randomly from the global view with the scale of (0.05, 0.6). Image patches cropped from the global view are resized to 64. The number of patch-based feature $L$ per anchor image is 5. SGD optimizer is used with a learning rate decays by cosine scheduler from 0.1 to 0 with batch size 256 on 2 Nvidia RTX 3090 in 200 epochs. For Places-LT, we only fine-tune the last block of the backbone for 30 epochs~\cite{kang2019decoupling}. At the second stage, the parameters are the same as~\cite{li2021targeted}. The linear classifier is trained for 40 epochs with CE loss and class-balanced sampling~\cite{kang2019decoupling} with batch size 2048 using SGD optimizer. The learning rate is initialized as 10, 30, 2.5 for ImageNet-LT, iNaturaList 2018, and Places-LT, respectively, and multiplied by 0.1 at epoch 20 and 30.
 
 \subsection{Ablation Study}\label{sec:ablation}
 \begin{table}
    \centering 
   \small
     \setlength{\tabcolsep}{6.0pt}
     \begin{tabular}{l|cccc}
     \hline
     Settings & Many & Medium & Few & Overall \\ \hline
     Baseline  & 61.6 & 48.6 & 30.3 & 51.2 \\
     DSCL  & 63.4 & 50.0 & 31.4 & 52.6 \\
     + PBSD & 67.2 & 53.7 & 33.7 & 56.3 \\
     DSCL + PBSD$^*$ & 67.2 & 53.9 & 33.7 & 56.2 \\
     DSCL + PBSD$\dagger$ & 68.0 & 53.3 & 32.3 & 56.1 \\
     DSCL + PBSD & \textbf{68.5} & \textbf{55.2} & \textbf{35.4} & \textbf{57.7} \\ \hline
     \end{tabular}
     \caption{Effectiveness of each component in our method on ImageNet-LT. SCL is used as baseline. * denotes using features of the global view instead of patch-based features to calculate Eq.~\eqref{eq:loss_pbsd}. $\dagger$ denotes using the multi-crop trick~\cite{caron2020unsupervised} instead of PBSD.}
     \label{tab:components}
 
 \end{table}
 
 \begin{table}
     \small
     \centering
     \setlength{\tabcolsep}{12.0pt}
     \begin{tabular}{l|cc}
     \hline
     Settings & ResNet50 & ResNeXt50 \\ \hline
     Baseline  & 51.2 & 51.8 \\
     DSCL  & 52.6 & 53.2 \\
     + PBSD & 56.3 & 57.7 \\
     DSCL + PBSD & 57.7 & 58.7 \\ \hline
     \end{tabular}
     \caption{Ablation study of each component in our method on different backbones.}
     \label{tab:backbone}
 \end{table}
 
 \begin{table*}
     \setlength{\tabcolsep}{4.5pt}
     \small
     \centering
     \begin{tabular}{l|c|cccc|cccc|c}
     \hline
     \multirow{2}{*}{Methods}     & \multirow{2}{*}{Reference} & \multicolumn{4}{c|}{ImageNet-LT} & \multicolumn{4}{c|}{iNaturaList 2018} & Places-LT \\ \cline{3-11}
                         &                   & Many & Medium & Few & Overall  & Many & Medium & Few & Overall & Overall \\ \hline
     CE & - & 64.0 & 33.8 & 5.8 & 41.6 & 72.2 & 63.0 & 57.2 & 61.7 & 30.2 \\
     Balanced & NeurIPS20 & 61.1 & 47.5 & 27.6 & 50.1 & - & - & - & - & 38.7 \\
      \hline
     cRT & ICLR20 & 58.8 & 44.4 & 26.1 & 47.3 & 69.0 & 66.0 & 63.2 & 65.2 & 36.7\\
     DisAlign & CVPR21 & 59.9 & 49.9 & 31.8 & 51.3 & - & - & - & 67.8 & 39.3 \\ \hline
     BatchFormer & CVPR22 & 61.4 & 47.8 & 33.6 & 51.1 & - & - & - & - & 38.2 \\ \hline
     KCL & ICLR20 & 61.8 & 49.4 & 30.9 & 51.5 & - & - & - & 68.6 & - \\
     PaCo$\ddagger$ & ICCV21 & 59.7 & 53.2 & 38.1 & 53.6 & 66.3 & 70.8 & 70.6 & 70.2 & 41.2 \\
     TSC & CVPR22 & 63.5 & 49.7 & 30.4 & 52.4 & 72.6 & 70.6 & 67.8 & 69.7 & - \\
     BCL & CVPR22 & - & - & - & 56.0 & - & - & - & 71.8 & - \\ \hline
     \rowcolor{gray!30} Our* & This paper & 67.2 & 54.8 & 38.7 & 57.4 & - & - & - & - & - \\
     \rowcolor{gray!30} Our & This paper & 68.5 & 55.2 & 35.4 & \textbf{57.7} & 74.2 & 72.9 & 70.3 & \textbf{72.0} & \textbf{42.4} \\ \hline
     RIDE & ICLR21 & - & - & - & 55.4 & 70.9 & 72.4 & 73.1 & 72.6 & 40.3 \\
     NCL$\dagger$ & CVPR22 & - & - & - & 59.5 & 72.7 & 75.6 & 74.5 & \textbf{74.9} & - \\
     SADE & NeurIPS22 &  - & - & - & - & - & - & - & 72.9 & 40.9 \\ \hline
     \rowcolor{gray!30} Our + RIDE & This paper & 70.1 & 57.5 & 37.7 & \textbf{59.7} & 76.2 & 75.7 & 73.6 & \textbf{74.9} & - \\ \hline
     \end{tabular}
     \caption{Comparison with recent methods on ImageNet-LT, iNaturaList2018, and Places-LT. CE denotes training the model with the cross entropy loss. $*$ denotes the learning rate at the second stage of our method is initialized as 2.5. $\ddagger$ denotes the model is trained without RandAug~\cite{cubuk2020randaugment} and with 200 epochs for fair comparison. $\dagger$ denotes the model is trained with RandAug and 400 epochs, which is a more expensive training setup than ours.}
     \label{tab:sota_imagenet}
 \end{table*}
 
 \textbf{Components analysis.} We analyze the effectiveness of each proposed component on ImageNet-LT in Table~\ref{tab:components}.
 SCL~\cite{Khosla2020} is used as baseline.
 Compared with the SCL baseline, DSCL improves the top-1 accuracy by 1.4\%. This result is already better than the recent contrastive learning based method TSC~\cite{li2021targeted}. Many methods for long-tailed classification could improve the performance of tail classes but sacrifice the head classes performance. Different from those works, PBSD improves the performance of both head and tail classes. The Table~\ref{tab:components} clearly indicates that, the combination of DSCL and PBSD achieves the best performance.
 The introduction of patch-based features are important to PBSD. We conduct experiment by using features of global view to calculate Eq.~\eqref{eq:loss_pbsd}. It decreases the overall accuracy by about 1.5\%.
 In addition, our method is also more effective than the multi-crop trick, \emph{i.e.}, it improves the overall accuracy by 1.6\% over the multi-crop trick.
 In summarize, each component in our method is effective in boosting the performance.
 
 \noindent\textbf{Components analysis on different backbones.} To validate that our method generalizes well on different backbones, we further conduct experiments using the ResNeXt50~\cite{xie2017aggregated} as backbone on ImageNet-LT. Results are summarized in Table~\ref{tab:backbone}, where our proposed components are also effective on ResNext50. Both DSCL and PBSD can bring performance improvement over the baseline. The combination of them achieves the best performance.
 
 \begin{figure}
    \centering
 \includegraphics[width=1\linewidth]{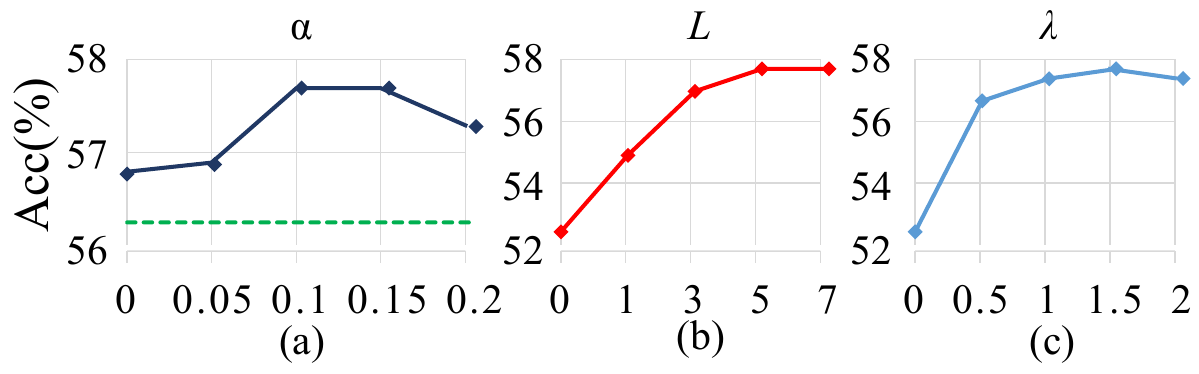}
 \caption{Evaluation of $\alpha$ in Eq.~\eqref{eq:decouple_target}, the number of patch-based features $L$ per anchor image, and the loss weight $\lambda$ on ImageNet-LT in (a), (b), and (c), respectively. Green dotted line in (a) denotes the baseline SCL.}
 \label{fig:parameters}
 \end{figure}
 
 \noindent\textbf{The impact of $\alpha$ in Eq.~\eqref{eq:decouple_target}} is investigated in Fig.~\ref{fig:parameters} (a). $\alpha$ determines the weight of pulling the anchor with its data augmented one. $\alpha=0$ means only pulling the anchor with other images from the same class. This setting decreases the accuracy from 57.7\% to 56.8\%, showing the importance of involving two kinds of positives. In addition, this setting still outperforms the SCL baseline, \emph{i.e.}, denoted by the green dotted line in the figure. It indicates that preventing the biased features is important. $\alpha=1$ degenerates the loss into the self-supervised loss. The accuracy is only 39.8\% because of the lack of label information. We set $\alpha$ as 0.1, which gets the best performance. Setting $\alpha=0.1$ also gets competitive performance on different datasets as shown in following experiments.
 
 \noindent\textbf{The impact of the number of patch-based features} per anchor image is shown in Fig.~\ref{fig:parameters} (b). The model benefits from involving more patch-based features into training. The top-1 accuracy improves from 55.0\% to 57.7\% when increasing $L$ from 1 to 5. We set $L$ as 5 for a reasonable trade-off between training cost and accuracy.
 
 \noindent\textbf{The impact of the loss weight $\lambda$} is shown in Fig.~\ref{fig:parameters} (c). Because $\lambda$ weights the influence of PBSD, the figure shows that PBSD is important. Setting $\lambda$ from 1 to 2 gets similar performance. We set it as 1.5 for different datasets.

 \subsection{Comparison with Recent Works}\label{sec:sota}
 
 We compare our method with recent works on ImageNet-LT, iNaturaList 2018, and Places-LT.
 The compared methods include re-balancing methods~\cite{ren2020balanced}, decoupling methods~\cite{kang2019decoupling,zhang2021distribution}, transfer learning based methods~\cite{hou2022batchformer}, methods that extend SCL~\cite{kang2020exploring,li2021targeted,cui2021parametric,zhu2022balanced}, and ensemble-based methods~\cite{li2022nested,zhang2022self,wang2020long}. Experimental results are summarized in Table~\ref{tab:sota_imagenet}.
 
 As shown in Table~\ref{tab:sota_imagenet}, directly using cross entropy loss leads to a poor performance on tail classes. Most long-tailed recognition methods improve the overall performance, but sacrifice the accuracy on `Many' split. Compared with the re-balancing methods, decoupling methods adjust the classifier after the training, and achieve a better performance, showing the effectiveness of the two-stage training strategy.
 Compared with above works, transfer learning based methods get better performance on head classes.
 For instance, BatchFormer gets a higher accuracy on `Many' split than DisAlign which has the same overall accuracy with it.
 
 Our method achieves the best overall accuracy of 57.7\% on ImageNet-LT. It also outperforms PaCo~\cite{cui2021parametric} that uses stronger data augmentation and twice training epochs. To make a fair comparison, we train PaCo with the same data augmentation and training epochs as our method, which decreases it accuracy from 57.0\% to 53.6\%.
 We also found that the learning rate of the second stage linear classifier training can change the accuracy distribution on `Many', `Medium' and `Few' splits, while maintaining the same overall accuracy. For instance, with a learning rate of 2.5 at the second training stage, the accuracy on `Few' split increases from 35.4\% to 38.7\%, while the overall accuracy only decreases by about 0.3\%. We hence note that, the overall accuracy could be more meaningful than individual accuracy on each split, which can be adjusted by hyperparameters.
 
 Our method can also be combined with ensemble-based method to further boost its performance. Combined with RIDE, our method achieves 59.7\% overall accuracy on ImageNet-LT, outperforming all those compared ensemble-based methods. Our method also achieves superior performance on iNaturaList 2018, where it gets comparable performance with NCL that is trained with stronger data augmentation and twice training epochs.
 With only a single model, our method achieves the best performance on Places-LT.
 
 \section{Conclusion}
 To tackle the challenge of long-tailed recognition, this paper analyzed two issues in SCL and addressed them with DSCL and PBSD. The DSCL decouples two types of positives in SCL, and optimizes their relations toward different objectives to alleviate the influence of the imbalanced dataset. The PBSD leverages head classes to facilitate the representation learning in tail classes by exploring patch-level similarity relationship. Experiments on different benchmarks demonstrated the promising performance of our method, where it outperforms recent works using more expensive setups. Extending our method to long-tailed detection is considered as the future work.

 \section*{Acknowledgements}
 This work is supported in part by Natural Science Foundation of China under Grant No. U20B2052, 61936011, in part by the Okawa Foundation Research Award.
 
 \bibliography{aaai24}

\begin{thebibliography}{34}
\providecommand{\natexlab}[1]{#1}

\bibitem[{Byrd and Lipton(2019)}]{byrd2019effect}
Byrd, J.; and Lipton, Z. 2019.
\newblock What is the effect of importance weighting in deep learning?
\newblock In \emph{ICML}, 872--881. PMLR.

\bibitem[{Caron et~al.(2020)Caron, Misra, Mairal, Goyal, Bojanowski, and
  Joulin}]{caron2020unsupervised}
Caron, M.; Misra, I.; Mairal, J.; Goyal, P.; Bojanowski, P.; and Joulin, A.
  2020.
\newblock Unsupervised learning of visual features by contrasting cluster
  assignments.
\newblock \emph{NeurIPS}, 33: 9912--9924.

\bibitem[{Chen et~al.(2020)Chen, Fan, Girshick, and He}]{Chen2020}
Chen, X.; Fan, H.; Girshick, R.; and He, K. 2020.
\newblock Improved baselines with momentum contrastive learning.
\newblock arXiv:2003.04297.

\bibitem[{Cubuk et~al.(2020)Cubuk, Zoph, Shlens, and Le}]{cubuk2020randaugment}
Cubuk, E.~D.; Zoph, B.; Shlens, J.; and Le, Q.~V. 2020.
\newblock Randaugment: Practical automated data augmentation with a reduced
  search space.
\newblock In \emph{CVPRW}, 702--703.

\bibitem[{Cui et~al.(2021)Cui, Zhong, Liu, Yu, and Jia}]{cui2021parametric}
Cui, J.; Zhong, Z.; Liu, S.; Yu, B.; and Jia, J. 2021.
\newblock Parametric contrastive learning.
\newblock In \emph{ICCV}, 715--724.

\bibitem[{He et~al.(2020)He, Fan, Wu, Xie, and Girshick}]{He2020}
He, K.; Fan, H.; Wu, Y.; Xie, S.; and Girshick, R. 2020.
\newblock Momentum contrast for unsupervised visual representation learning.
\newblock In \emph{CVPR}, 9729--9738.

\bibitem[{He et~al.(2017)He, Gkioxari, Doll{\'a}r, and Girshick}]{he2017mask}
He, K.; Gkioxari, G.; Doll{\'a}r, P.; and Girshick, R. 2017.
\newblock Mask r-cnn.
\newblock In \emph{ICCV}, 2961--2969.

\bibitem[{He et~al.(2016)He, Zhang, Ren, and Sun}]{He2016}
He, K.; Zhang, X.; Ren, S.; and Sun, J. 2016.
\newblock Deep residual learning for image recognition.
\newblock In \emph{CVPR}, 770--778.

\bibitem[{Hou, Yu, and Tao(2022)}]{hou2022batchformer}
Hou, Z.; Yu, B.; and Tao, D. 2022.
\newblock BatchFormer: Learning to Explore Sample Relationships for Robust
  Representation Learning.
\newblock arXiv:2203.01522.

\bibitem[{Japkowicz and Stephen(2002)}]{japkowicz2002class}
Japkowicz, N.; and Stephen, S. 2002.
\newblock The class imbalance problem: A systematic study.
\newblock \emph{Intelligent data analysis}, 6(5): 429--449.

\bibitem[{Kang et~al.(2020)Kang, Li, Xie, Yuan, and Feng}]{kang2020exploring}
Kang, B.; Li, Y.; Xie, S.; Yuan, Z.; and Feng, J. 2020.
\newblock Exploring balanced feature spaces for representation learning.
\newblock In \emph{ICLR}.

\bibitem[{Kang et~al.(2019)Kang, Xie, Rohrbach, Yan, Gordo, Feng, and
  Kalantidis}]{kang2019decoupling}
Kang, B.; Xie, S.; Rohrbach, M.; Yan, Z.; Gordo, A.; Feng, J.; and Kalantidis,
  Y. 2019.
\newblock Decoupling representation and classifier for long-tailed recognition.
\newblock arXiv:1910.09217.

\bibitem[{Khosla et~al.(2020)Khosla, Teterwak, Wang, Sarna, Tian, Isola,
  Maschinot, Liu, and Krishnan}]{Khosla2020}
Khosla, P.; Teterwak, P.; Wang, C.; Sarna, A.; Tian, Y.; Isola, P.; Maschinot,
  A.; Liu, C.; and Krishnan, D. 2020.
\newblock Supervised contrastive learning.
\newblock arXiv:2004.11362.

\bibitem[{Li et~al.(2022)Li, Tan, Wan, Lei, and Guo}]{li2022nested}
Li, J.; Tan, Z.; Wan, J.; Lei, Z.; and Guo, G. 2022.
\newblock Nested Collaborative Learning for Long-Tailed Visual Recognition.
\newblock In \emph{CVPR}, 6949--6958.

\bibitem[{Li et~al.(2021)Li, Cao, Yuan, Fan, Yang, Feris, Indyk, and
  Katabi}]{li2021targeted}
Li, T.; Cao, P.; Yuan, Y.; Fan, L.; Yang, Y.; Feris, R.; Indyk, P.; and Katabi,
  D. 2021.
\newblock Targeted Supervised Contrastive Learning for Long-Tailed Recognition.
\newblock arXiv:2111.13998.

\bibitem[{Liu et~al.(2019)Liu, Miao, Zhan, Wang, Gong, and Yu}]{liu2019large}
Liu, Z.; Miao, Z.; Zhan, X.; Wang, J.; Gong, B.; and Yu, S.~X. 2019.
\newblock Large-scale long-tailed recognition in an open world.
\newblock In \emph{CVPR}, 2537--2546.

\bibitem[{Long, Shelhamer, and Darrell(2015)}]{long2015fully}
Long, J.; Shelhamer, E.; and Darrell, T. 2015.
\newblock Fully convolutional networks for semantic segmentation.
\newblock In \emph{CVPR}, 3431--3440.

\bibitem[{Ma et~al.(2023)Ma, Jiao, Liu, Yang, Liu, and Li}]{ma2023curvature}
Ma, Y.; Jiao, L.; Liu, F.; Yang, S.; Liu, X.; and Li, L. 2023.
\newblock Curvature-Balanced Feature Manifold Learning for Long-Tailed
  Classification.
\newblock In \emph{CVPR}.

\bibitem[{Quan et~al.(2019)Quan, Dong, Wu, Zhu, and Yang}]{quan2019auto}
Quan, R.; Dong, X.; Wu, Y.; Zhu, L.; and Yang, Y. 2019.
\newblock Auto-reid: Searching for a part-aware convnet for person
  re-identification.
\newblock In \emph{ICCV}, 3750--3759.

\bibitem[{Ren et~al.(2020)Ren, Yu, Ma, Zhao, Yi et~al.}]{ren2020balanced}
Ren, J.; Yu, C.; Ma, X.; Zhao, H.; Yi, S.; et~al. 2020.
\newblock Balanced meta-softmax for long-tailed visual recognition.
\newblock \emph{NeurIPS}, 33: 4175--4186.

\bibitem[{Russakovsky et~al.(2015)Russakovsky, Deng, Su, Krause, Satheesh, Ma,
  Huang, Karpathy, Khosla, Bernstein et~al.}]{Russakovsky2015}
Russakovsky, O.; Deng, J.; Su, H.; Krause, J.; Satheesh, S.; Ma, S.; Huang, Z.;
  Karpathy, A.; Khosla, A.; Bernstein, M.; et~al. 2015.
\newblock Imagenet large scale visual recognition challenge.
\newblock \emph{IJCV}, 115(3): 211--252.

\bibitem[{Sun et~al.(2018)Sun, Zheng, Yang, Tian, and Wang}]{sun2018beyond}
Sun, Y.; Zheng, L.; Yang, Y.; Tian, Q.; and Wang, S. 2018.
\newblock Beyond part models: Person retrieval with refined part pooling (and a
  strong convolutional baseline).
\newblock In \emph{ECCV}, 480--496.

\bibitem[{Van~Horn et~al.(2018)Van~Horn, Mac~Aodha, Song, Cui, Sun, Shepard,
  Adam, Perona, and Belongie}]{van2018inaturalist}
Van~Horn, G.; Mac~Aodha, O.; Song, Y.; Cui, Y.; Sun, C.; Shepard, A.; Adam, H.;
  Perona, P.; and Belongie, S. 2018.
\newblock The inaturalist species classification and detection dataset.
\newblock In \emph{CVPR}, 8769--8778.

\bibitem[{Vaswani et~al.(2017)Vaswani, Shazeer, Parmar, Uszkoreit, Jones,
  Gomez, Kaiser, and Polosukhin}]{vaswani2017attention}
Vaswani, A.; Shazeer, N.; Parmar, N.; Uszkoreit, J.; Jones, L.; Gomez, A.~N.;
  Kaiser, {\L}.; and Polosukhin, I. 2017.
\newblock Attention is all you need.
\newblock \emph{NeurIPS}, 30.

\bibitem[{Wang et~al.(2020)Wang, Lian, Miao, Liu, and Yu}]{wang2020long}
Wang, X.; Lian, L.; Miao, Z.; Liu, Z.; and Yu, S.~X. 2020.
\newblock Long-tailed recognition by routing diverse distribution-aware
  experts.
\newblock arXiv:2010.01809.

\bibitem[{Wei et~al.(2020)Wei, Xie, He, Chang, Zhang, Zhou, Li, and
  Tian}]{wei2020can}
Wei, L.; Xie, L.; He, J.; Chang, J.; Zhang, X.; Zhou, W.; Li, H.; and Tian, Q.
  2020.
\newblock Can Semantic Labels Assist Self-Supervised Visual Representation
  Learning?
\newblock arXiv:2011.08621.

\bibitem[{Xie et~al.(2017)Xie, Girshick, Doll{\'a}r, Tu, and
  He}]{xie2017aggregated}
Xie, S.; Girshick, R.; Doll{\'a}r, P.; Tu, Z.; and He, K. 2017.
\newblock Aggregated residual transformations for deep neural networks.
\newblock In \emph{CVPR}, 1492--1500.

\bibitem[{Yun et~al.(2022)Yun, Lee, Kim, and Shin}]{yun2022patch}
Yun, S.; Lee, H.; Kim, J.; and Shin, J. 2022.
\newblock Patch-level representation learning for self-supervised vision
  transformers.
\newblock In \emph{CVPR}, 8354--8363.

\bibitem[{Zhang et~al.(2014)Zhang, Donahue, Girshick, and
  Darrell}]{zhang2014part}
Zhang, N.; Donahue, J.; Girshick, R.; and Darrell, T. 2014.
\newblock Part-based R-CNNs for fine-grained category detection.
\newblock In \emph{ECCV}, 834--849. Springer.

\bibitem[{Zhang et~al.(2021)Zhang, Li, Yan, He, and
  Sun}]{zhang2021distribution}
Zhang, S.; Li, Z.; Yan, S.; He, X.; and Sun, J. 2021.
\newblock Distribution alignment: A unified framework for long-tail visual
  recognition.
\newblock In \emph{CVPR}, 2361--2370.

\bibitem[{Zhang et~al.(2023)Zhang, Zhou, Wang, Wang, and Yan}]{zhang2023patch}
Zhang, S.; Zhou, Q.; Wang, Z.; Wang, F.; and Yan, J. 2023.
\newblock Patch-level Contrastive Learning via Positional Query for Visual
  Pre-training.
\newblock In \emph{ICML}, 41990--41999. PMLR.

\bibitem[{Zhang et~al.(2022)Zhang, Hooi, Hong, and Feng}]{zhang2022self}
Zhang, Y.; Hooi, B.; Hong, L.; and Feng, J. 2022.
\newblock Self-supervised aggregation of diverse experts for test-agnostic
  long-tailed recognition.
\newblock \emph{NeurIPS}, 35: 34077--34090.

\bibitem[{Zhou et~al.(2017)Zhou, Lapedriza, Khosla, Oliva, and
  Torralba}]{zhou2017places}
Zhou, B.; Lapedriza, A.; Khosla, A.; Oliva, A.; and Torralba, A. 2017.
\newblock Places: A 10 million image database for scene recognition.
\newblock \emph{TPAMI}, 40(6): 1452--1464.

\bibitem[{Zhu et~al.(2022)Zhu, Wang, Chen, Chen, and Jiang}]{zhu2022balanced}
Zhu, J.; Wang, Z.; Chen, J.; Chen, Y.-P.~P.; and Jiang, Y.-G. 2022.
\newblock Balanced contrastive learning for long-tailed visual recognition.
\newblock In \emph{CVPR}, 6908--6917.

\end{thebibliography}

\end{document}